\documentclass[subscriptcorrection,upint,varvw,barcolor=Goldenrod3,mathalfa=cal=euler,balance,hyphenate,french,pdf-a,nolists,nocopyright]{asmejour} %
\usepackage{algorithmic}
\usepackage{graphicx}
\usepackage{textcomp}
\usepackage{xcolor}
\usepackage{mathtools}
\usepackage{array}
\usepackage{tabularx}
\usepackage{ragged2e}
\usepackage{booktabs}
\usepackage{titlesec}
\usepackage{siunitx}
\usepackage{makecell}
\usepackage{changes}

\hypersetup{%
	pdfauthor={Maryam Seraj},                       		   	
	pdftitle={Optimal Design of a Cable-Driven Coaxial Spherical Parallel Mechanism (CDC-SPM) for Medical Tasks},                  	
	pdfkeywords={Spherical Parallel Mechanism (SPM), Remote  Center  of  Motion  (RCM), Kinematics, Haptic Interface},
	pdfsubject = {Cable-Driven Coaxial Spherical Parallel Mechanism (CDC-SPM)},			
}

\JourName{Mechanical Design}

\begin{document}



\SetAuthorBlock{Maryam Seraj\CorrespondingAuthor, Mohammad H. Kamrava, Carlo Tiseo\CorrespondingAuthor}{Engineering/Informatics department,\\
  University of Sussex,\\
  Richmond building, Falmer \\
   Brighton, BN19QT, United Kingdom \\
   email: m.seraj@sussex.ac.uk, mhossein.kamrava@gmail.com, c.tiseo@sussex.ac.uk
} 



\title{Parametric Design of a Cable-Driven Coaxial Spherical Parallel Mechanism for Ultrasound Scans}


\begin{abstract}
Haptic interfaces play a critical role in medical teleoperation by enabling surgeons to interact with remote environments through realistic force and motion feedback. Achieving high fidelity in such systems requires balancing the trade-offs among workspace, dexterity, stiffness, inertia, and bandwidth, particularly in applications demanding pure rotational motion. This paper presents the design methodology and kinematic analysis of a Cable-Driven Coaxial Spherical Parallel Mechanism (CDC-SPM) developed to address these challenges. The proposed approach focuses on the mechanical design and parametric synthesis of the mechanism to meet task-specific requirements in medical applications. In particular, the design enables the relocation of the center of rotation to an external point corresponding to the tool–tissue interaction, while ensuring appropriate workspace coverage and collision avoidance. The proposed cable-driven interface design allows for reducing the mass placed at the robot arm end-effector, thereby minimizing inertial loads, enhancing stiffness, and improving dynamic responsiveness. Through parallel and coaxial actuation, the mechanism achieves decoupled rotational degrees of freedom with isotropic force and torque transmission. A prototype is developed to validate the mechanical feasibility and kinematic behavior of the proposed mechanism. These results demonstrate the suitability of the proposed mechanism design for future integration into haptic interfaces for medical applications such as ultrasound imaging.
\end{abstract}

\date{Version \versionno, \today}

\maketitle 


\section{Introduction}

Achieving human-like dexterity in robots is a key goal for manufacturers and service industries, including medical applications. Human dexterity arises from the seamless integration of sensing, motion control, and adaptive compliance, enabling precise interaction with complex and unstructured environments \cite{Huang2025Human-likeReview}. The replication of such skills in robots is crucial for their ability to handle delicate tasks, such as human-robot interaction, safely and effectively. For example, in medicine, robotic systems capable of mimicking human-like dexterity can improve surgical precision, reduce fatigue for clinicians, and increase access to skilled care through remote operation \cite{Morgan2022RobotsReview}. Without these abilities, robots remain limited to performing rigid, predefined motions and cannot achieve the fine dynamic interaction control needed for dynamic procedures, such as palpation, ultrasound imaging, or rehabilitation therapies.

Robot arms often struggle to replicate human-like movements and manipulation \cite{Tiseo2022RobustIK-Optimisation,Tiseo2021HapFIC:Systems, Tiseo2022AchievingRobotics,Tiseo2024SafeControllers}. Robotic hands might be a solution for general-purpose applications, but specialized end-effector tools mimicking the role of the human hand are better suited for robots that are used for specific applications \cite{Tiseo2022AchievingRobotics}. For example, robots used for medical applications are required at least to match the clinician's performance, both in the case of teleoperation and manipulation. The gap between robot and human performance is mainly related to high system inertia, limited actuation bandwidth, and unaddressed dynamic effects \cite{Mehrdad2020ReviewTelerobots}. These limitations cause robotic systems to often respond more slowly and less precisely than human during interaction with the environment \cite{Tiseo2022RobustIK-Optimisation}. High inertia reduces responsiveness and makes it difficult for the robot to perform quick and delicate adjustments, while limited bandwidth constrains the range of achievable forces and motions, reducing teleoperation intuitiveness \cite{Babarahmati2019FractalRobotics}. Meanwhile, unaddressed dynamic effects like vibration, friction, and cable elasticity can distort force feedback. These effects can compromise the accuracy and stability of control, especially in tasks in which continuous contact with soft or moving tissue is needed \cite{Riener2023DoDomains}, \cite{Genaldy1990ATasks}, which is crucial in applications such as rehabilitation and ultrasound scans \cite{Tiseo2022AchievingRobotics}. In contrast, they are less critical in laparoscopic surgeries or position control surgeries (e.g., orthopedics) where a rigid and precise control of the tool is required \cite{Li2023RoboticReview, Alleblas2016ErgonomicsPerspective, Zhou2011EffectAcquisition}.

To address these issues, haptic interfaces have been designed to bridge the gap between human intent and robotic execution by allowing operators to control robots more naturally and intuitively. For example, in a teleoperation system, a haptic device acts as the physical link between the operator and the robot, allowing them to interact through forces and motions in real time \cite{Payne2021Shared-ControlRobots}. This two-way exchange of mechanical information lets the operator feel and control the interaction as if they were directly moving the object. This is essential for achieving precision, safety, and natural responsiveness in medical teleoperation tasks \cite{Hayward1996PerformanceInterfaces}.

\begin{figure}[htbp]
\centering
\includegraphics[width=0.97\columnwidth]{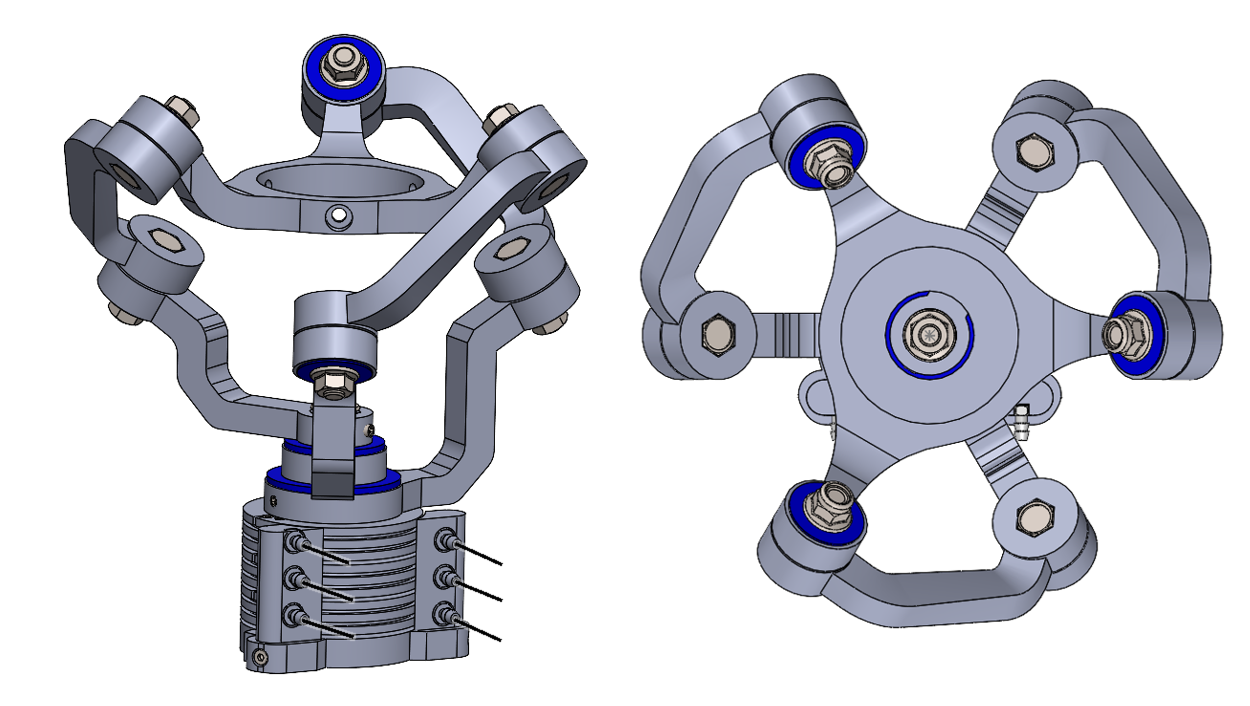}
\caption{Schematic representation of the proposed cable-driven coaxial spherical parallel haptic interface.}
\label{fig:ProposedModel}
\end{figure}

However, achieving high-quality haptic interaction depends not only on control performance, but also on the mechanical characteristics of the underlying system. In particular, mechanisms intended for haptic interaction must satisfy several key requirements such as low projected inertia, high structural stiffness, accurate alignment of the center of rotation with the interaction point, and smooth, singularity-free motion \cite{Hannaford2008Haptics}. These requirements are particularly critical in applications such as robot-assisted ultrasound, where both dexterity and stable force regulation are required \cite{Chen2024DesignImaging}.

In the context of medical robotics, various robotic systems have been developed for ultrasound examinations, including teleoperated and semi-autonomous platforms \cite{Salcudean1999Robot-AssistedExperiments}, \cite{Nouaille2012ProcessRobot}. More recent approaches have explored collaborative and industrial robotic arms to improve safety and accessibility \cite{Mathiassen2016AnUR5}. While these systems enhance repeatability and reduce operator workload, they often rely on serial or hybrid architectures, which can exhibit relatively high inertia and limited capability to enforce a precise remote center of motion during probe manipulation \cite{Huang2023ReviewApplications}, \cite{Jiang2023RoboticPerspectives}. In parallel, the design of robotic mechanisms for ultrasound has been extensively investigated, with particular attention to stiffness, dexterity, and force control requirements \cite{Du2024ATechnology}. Despite these advances, most existing systems still rely on such serial or hybrid architectures, and therefore inherit similar limitations.

From a kinematic perspective, spherical parallel mechanisms (SPMs) offer a promising alternative for applications requiring pure rotational motion about a fixed center. These mechanisms provide high stiffness and compactness \cite{Saafi2020ForwardValidation}, and recent developments such as coaxial-input architectures improve motion continuity and reduce singularities \cite{Tursynbek2021InfiniteAxes}. However, in most existing designs, the center of rotation is located within the mechanism structure, which prevents alignment with the tool–patient contact point. As a result, pure rotation about the interaction point cannot be achieved, which limits its suitability for applications involving direct physical interaction, such as ultrasound imaging. Therefore, there remains a need for a mechanism that can both preserve the advantages of spherical parallel architectures and satisfy the mechanical requirements of haptic interaction. In particular, enabling rotation about an externally defined point, corresponding to the tool–tissue contact location, is essential for improving interaction fidelity and usability.

This paper proposes a cable-driven coaxial spherical parallel mechanism (CDC-SPM), designed to meet key requirements of haptic interaction in medical applications. The proposed mechanism: (i) relocates the center of rotation to an external point corresponding to the tool–patient contact location, (ii) enables parametric design framework that adapts the mechanism geometry to the dimensions of the target medical tool, and (iii) employs cable-driven actuation with coaxial input axes to reduce reflected inertia and improve haptic transparency. The mechanism is formulated, analyzed, and experimentally validated, demonstrating its suitability as the core component of a haptic interface. 

The remainder of the paper is structured as follows. First, key design principles for haptic interfaces are reviewed in the context of medical applications. Next, existing mechanism architectures are analyzed, highlighting the limitations of conventional spherical parallel manipulators. Based on these limitations, the CDC-SPM is introduced with its structure, kinematics, and mechanical design. This is followed by parametric design, kinematic analysis, workspace evaluation, and the development of both the CAD model and physical prototype. Finally, the mechanism’s suitability for ultrasound imaging is evaluated, and possible future improvements are discussed.

\section{Preliminaries on Design Requirements for Haptic Interaction}
Haptic interaction relies on mechanisms that enable a safe, intuitive, and effective exchange of motion and force between the operator and the environment. From a design perspective, these capabilities are not only determined by control strategies, but are fundamentally constrained by the mechanical characteristics of the underlying system. Therefore, the following criteria have been identified to determine how well the device can replicate realistic touch, accommodate operator needs, and support specific application requirements \cite{Torabi2021KinematicReview}. The workspace should be designed to ensure that the interface retains an effective manipulability and, consequently, dexterity within the task workspace. The stiffness and effort (i.e., forces and torques) ranges define the ability of the interface to match environmental dynamics. Weight, friction, and back-drivability determine the intrinsic impedance of the interface. Bandwidth determines the ability of the interface to adapt to changes in environmental interaction. Consequently, for safe and effective operation, must closely match the operator’s needs and requirements of the specific task \cite{Torabi2021KinematicReview}. A well-designed haptic system enables precise and reliable transmission of force and motion, supporting high-fidelity, safe contact with the environment. In contrast, a haptic device that is not capable of delivering the aforementioned characteristics can lead to reduced performance, operator fatigue, and even potential safety risks in clinical practice \cite{PhilipKortum2008HCIInterfaces}. Recent research also highlights the interdependence of haptic realism, low inertia, high stiffness, force transparency, and ergonomic usability \cite{Torabi2021KinematicReview}. The delicate balance of these requirements necessitates thoughtful mechanical and control system design, especially as clinical applications become more demanding and diverse.  

\subsection{Mechanism Design for Medical Interaction Interfaces}
The design of a mechanism for medical applications starts with the definition of its structural configuration, including the required Degrees of Freedom (DoF), mechanism type, and the arrangement of links, joints, and actuators to deliver the desired motion and force feedback. Over the years, developers have created haptic interfaces for many applications, from simple single-DoF devices to complex multi-DoF systems \cite{Hayward1996PerformanceInterfaces}. In general, increasing the number of DoF can expand the device’s workspace compared to lower-DoF systems of similar size \cite{PhilipKortum2008HCIInterfaces}. The final choice of mechanism is also guided by the specific application and the part of the human body that interacts with the device, ensuring optimal comfort and control. From a mechanical design perspective, parallel mechanisms are often preferred due to their high structural stiffness, low inertia, and favourable force transmission characteristics, which are critical for interaction tasks \cite{Xia2025AdvancesDesigns}. As a result, most mechanisms used in haptic and interaction systems adopt parallel architectures, such as Delta.3, Omega.x, and Sigma.7 (Force Dimension, Switzerland), \cite{ForceDelta.3, ForceOmega.x, ForceSigma.7}, as well as Stewart platforms \cite{Le2022ApplicationRobot}, or SPMs \cite{Birglen2002SHaDeDevice}, \cite{Saafi2020ForwardValidation}. These designs provide high stiffness, low dynamic coupling, and precise motion transmission for both translational and rotational tasks. 

 Parallel mechanisms are commonly arranged in a symmetric structure that defines a clear separation between the fixed base and the moving platform \cite{Taghirad2013ParallelControl}. In architectures designed for rotational motion, such as spherical parallel mechanisms (see Fig.~\ref{fig:SPMandCSPM}a), this symmetry often results in a distinct center of rotation (CoR) where the joint axes intersect. While this characteristic is advantageous for achieving pure rotational motion, the location of the CoR is inherently fixed by the mechanism geometry. To date, most SPMs developed for medical applications have been optimized for scenarios where either the displacement at the instrument tip was negligible \cite{Stoianovici2013EndocavityManipulators} or the structural weight was not a primary concern \cite{Courreges2004AdvancesSystem}. Despite the advantages of these geometric arrangements in terms of stability and separation between translation and rotation, these characteristics do not necessarily align with the requirements of physical interaction tasks in medical applications. For example, when the robotic arm is used to hold or manipulate tools in procedures such as Minimally Invasive Surgery (MIS) or Ultrasound Scanning (US), it must ensure a higher level of precision and safety. This requires the mechanical center of rotation to coincide with the tip of a medical instrument or probe. When the CoR is misaligned with the tool tip, pure rotational motion about the interaction point cannot be achieved, leading to reduced dexterity and increased risk of tissue damage \cite{Yang2025SensorlessEstimation}, \cite{Aflatooni2023AlignmentArthroplasty}. Studies in robotic ultrasound have shown that properly maintaining the probe orientation at a point of contact is challenging because small off-axis rotations significantly compromise image stability and diagnostic quality \cite{Ma2022A-SEE:Applications}. Similarly, remote center of motion (RCM) mechanisms for MIS are purposely configured to limit the rotation about the entry point, which serves to protect tissue integrity and improve operative ease \cite{Zhou2018NewSurgery}, \cite{Beira2011Dionis:Surgery}. Despite these developments, existing spherical and parallel mechanisms do not simultaneously address the combined requirements of low inertia, precise alignment of the center of rotation with an externally defined interaction point, and smooth, continuous motion. This limitation highlights the need for novel mechanism designs for medical interaction applications.

\subsection{Task Requirements for Ultrasound Scan}
\begin{figure}[htbp]
\centerline{\includegraphics[width=.55\columnwidth]{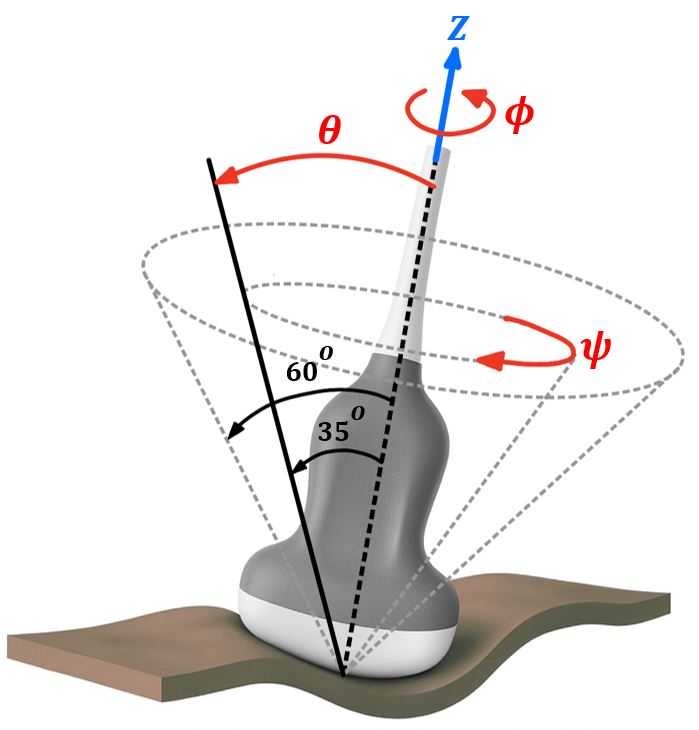}}
\caption{Probe workspace cone: useful workspace and safety region reported in \cite{Essomba2012DesignSystem}.}
\label{fig:UltrasoundProbeWorkspace}
\end{figure}

Ultrasound procedures require the probe to pivot smoothly about a fixed contact point on the patient’s skin, with controlled angular motion to maintain safety and image quality. Clinical studies show that in order to obtain high-quality images during ultrasound scanning, the probe should be angled below $35~\unit{\deg}$ (useful workspace) and never exceed $60~\unit{\deg}$ to $75~\unit{\deg}$ (safety region) to avoid patient discomfort or unintentional collisions (see Fig.~\ref{fig:UltrasoundProbeWorkspace}) \cite{Essomba2012DesignSystem}. The probe must keep stable contact, allow rotation about its axis for fine adjustments, and restrict unwanted translation \cite{Fenster19963-DReview}. These demands highlight the need for three rotational degrees of freedom about a well-defined fixed center of rotation at the point of contact.

The proposed CDC-SPM architecture is designed to satisfy these constraints. By relocating the center of rotation to the probe tip, the mechanism enables rotation about the contact point while minimizing unintended translational motion. Its cable-driven actuation and coaxial arrangement also aim to support low-inertia motion and ergonomic alignment, which are essential for developing a device suitable for ultrasound imaging tasks.

\section{Proposed CDC-SPM Haptic Interface}
As mentioned earlier, in certain medical applications like minimally invasive surgery or diagnostic ultrasound, the instrument tip needs to rotate around a fixed point. This requires pure, stable rotational motion that is precisely centered at the tool-patient interface. Achieving this demands highly specialized mechanics and kinematics in the interface device. Spherical Parallel Mechanisms (SPMs) are a proven solution for achieving pure rotational motion with parallel structure. This makes them highly attractive in haptic interface design, where large workspace, high structural stiffness, and low inertia are essential.

In addition, when a haptic interface is mounted on the end-effector of a robotic arm, as in targeted applications, several structural modifications are required to improve dexterity. To minimize displacement errors in tasks demanding pure rotational motion,the CoR should be directly located at the instrument tip, rather than relying on effective arm compensation to decouple translation and rotation. Moreover, reducing the inertia of the interface is essential to achieve high responsiveness and transparency. However, most existing designs place the CoR within the mechanism structure itself (either below or above the moving platform), rather than at the clinically relevant interaction point \cite{Zhang2024StateRobotics}.

\subsection{Structural Concept}
\begin{figure}[htbp]
\centerline{\includegraphics[width=\columnwidth]{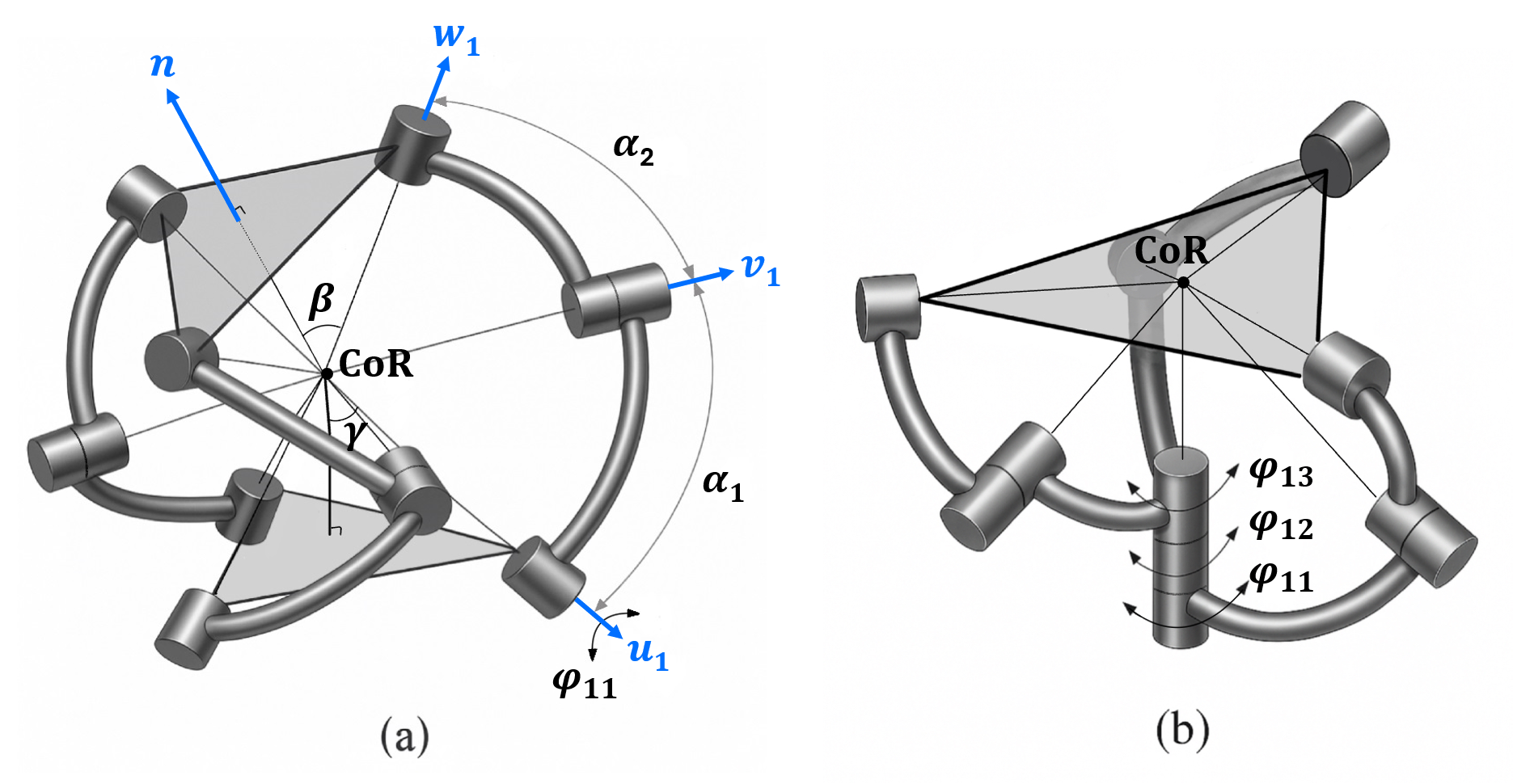}}
\caption{Geometry of a special spherical parallel manipulator: (a) General model; (b) Coaxial model with $\gamma = 0$ \cite{Bai2009ModellingParameters}.}
\label{fig:SPMandCSPM}
\end{figure}

A classic SPM, as illustrated in Fig.~\ref{fig:SPMandCSPM}a, consists of a fixed base, a moving platform, and three identical kinematic chains numbered as $~i\in \{1,2,3\}$ in the counterclockwise direction. Each of these chains is composed of two curved links, proximal (lower) and distal (upper), and a revolute-revolute-revolute joint configuration \cite{He2021AStroke}. Only the first revolute joint is actuated, and the remaining joints are passive. The unit vectors along the axes of the actuators are denoted by $\mathbf{u}_i$, while the unit vectors along the axes of the joints attached to the platform are denoted by $\mathbf{v}_i$. Finally, the unit vectors defined along the axes of the intermediate joints are noted as $\mathbf{w}_i$. Moreover, $\mathbf{n}_i$ represents the normal vector of the moving platform. The angles $\alpha_1$ and $\alpha_2$ define the curvature of the proximal and distal links, respectively. The angles $\beta$ and $\gamma$ define the geometry of the regular pyramid forming the mobile platform. 

In this conventional architecture, the axes of all joints, both active and passive, intersect at a single point that serves as the center of rotation (CoR) for the end-effector \cite{Bai2009ModellingParameters}. This configuration ensures that the structure provides a pure rotational movement around the CoR, while this specific point remains fixed with respect to the base \cite{Gosselin1993OnManipulators}, \cite{Gosselin1995AManipulators}. It also divides the robot into two symmetrical pyramids, the moving and fixed platforms. This setup is particularly useful for haptic applications since it reduces dynamic coupling and interference, allowing for a large, usable workspace. It allows the device to provide high acceleration, stiffness, and bandwidth, which are essential for precise and responsive force feedback \cite{Asada1985KinematicDesign}. By further aligning the rotational axes of the base and actuators so that the input axes are coaxial ($\gamma = 0$), the lower pyramid structure disappears, resulting in a more compact SPM that is both efficient and mechanically robust (see Fig.~\ref{fig:SPMandCSPM}b). This design improvement reduces misalignment, reduces friction, and removes workspace singularities that often affect performance in traditional designs. The coaxial arrangement enables a more streamlined form factor and improves force transmission. These factors are becoming more important in applications that require precise, real-time control, such as medical probe manipulation.

\begin{figure*}[!t]
\centerline{\includegraphics[width=1.0\linewidth]{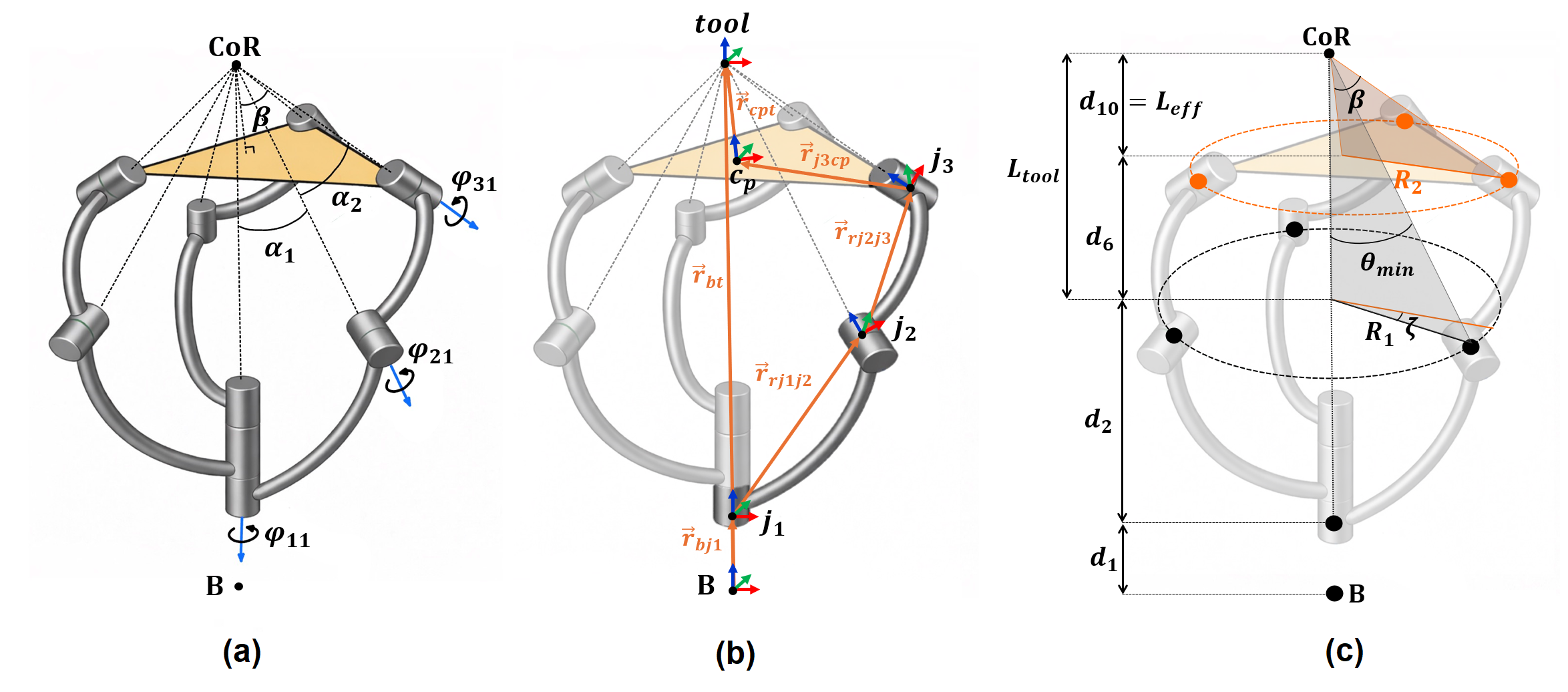}}
\caption{Modified CSPM: (a) Geometry, (b) Closed-loop kinematic chain, (c) Parametric design for task-oriented medical applications.}
\label{fig:ModifiedCSPM}
\end{figure*}

However, for medical applications, the CoR of the manipulator is usually located within or near the moving platform, rather than at the tool–patient contact point (the probe tip). This mismatch limits ergonomic accuracy and realistic motion reproduction. To address this limitation, this work introduces the Cable-Driven Coaxial Spherical Parallel Mechanism (CDC-SPM). Unlike most existing SPMs that place the center of rotation within or on the moving platform (see Fig.~\ref{fig:SPMandCSPM}), the proposed structure relocates the CoR to a point above the moving platform, right at the tip of the medical instrument (see Fig.~\ref{fig:ModifiedCSPM}a). This configuration aims to enable pure rotational motion at the probe tip while supporting a compact and miniaturized device layout. Since parameters such as inertia, stiffness, and workspace are interdependent, the mechanism is designed to explore a balance among these properties that is suitable for achieving high-quality haptic performance. In the proposed CDC-SPM, heavy actuators are connected to the active joints using cable-driven transmission and Bowden cables, relieving the robot arm and the haptic interface from the majority of their inertia. Despite the rotor and transmission rotational inertia still being transmitted to the joints, such a configuration significantly reduces device inertia, thereby improving responsiveness and haptic rendering speed  \cite{Qian2018ARobots}, \cite{LetierBowdenApplications}. Meanwhile, the parallel structure retains all of the key advantages of SPMs: It offers high stiffness for force accuracy and transparent force reflection for intuitive clinical use. It also provides ergonomic alignment for safe and comfortable operation during repetitive or long-term use. 

The proposed CDC-SPM is intended to support tasks that require natural, skillful rotation about a fixed point, such as real-time medical imaging or instrument guidance, by providing a mechanism whose architecture is designed to balance these interacting performance factors.

\subsection{Proposed Mechanism Kinematics}
Once the geometric constraints of the manipulator are defined, a parametric model of the interface kinematics can be formulated. The first geometric parameter to be defined is the total displacement from the $base$ to the $tool$ tip. The closed-loop vector chain for each leg is formulated as illustrated in Fig.~\ref{fig:ModifiedCSPM}b:

{\small
\begin{equation}
\vec{r}_{bt} = \vec{r}_{bj_{1i}} + \vec{r}_{j_1j_2} + \vec{r}_{j_2j_3} + \vec{r}_{j_3cp} + \vec{r}_{cpt}, \quad \forall ~i \in \{1,2,3\}.
\label{eq:ClosedLoopVector}
\end{equation}
}

This vector chain expresses the sequential positions of key points along the leg, starting from the base frame $B$ and ending at the tool tip $tool$. It provides the geometric foundation (the positions and orientations of each link's reference) for subsequent forward and inverse kinematic computations. 

\subsubsection{Frame Definitions}
To standardize the assignment of coordinate frames along the robotic structure, the Denavit–Hartenberg (DH) parameterization is employed. By extracting the DH parameters from the vector chain in Eq.~\ref{eq:ClosedLoopVector}, the representation of link transformations is simplified, ensuring consistency and reliability in kinematic modeling.

To explicitly relate the geometric design parameters to the Denavit–Hartenberg model, the key inter-joint vectors are computed from the manipulator geometry. 
The displacement from the $base$ to the first joint is: 
\[
\vec{r}_{b_{j1}}=
\begin{bmatrix}
0\\
0\\
d_1
\end{bmatrix},
\quad \forall ~i\in \{1,2,3\}.
\]
The radii $R_1$ and $R_2$ denote, respectively, the radius of the circle on which all second joints lie and all third joints lie. The position of $joint~2$ relative to $joint~1$ is obtained from the curvature angle $\alpha_1$ as:
\[
\vec{r}_{j_1j_2}=
\begin{bmatrix}
R_1 \\
0 \\
-(R_1 - z_{CoR}\tan\alpha_1)/\tan\alpha_1-d_1
\end{bmatrix},
\]
which directly yields the DH parameters $a_2,a_3,d_2$.
The second curvature angle $\alpha_2$ and the center of rotation determine the location of $joint~3$,  so the position of $joint~3$ relative to $joint~2$ is: 
\[
\vec{r}_{j_2j_3} =
\begin{bmatrix}
R_2\cos(\xi) - R_1 \\
R_2\sin(\xi) \\
z_{CoR}-L_{\mathrm{eff}}+({R_1 - z_{CoR}\tan(\alpha_1)})/{\tan(\alpha_1)}
\end{bmatrix},
\]
whose components correspond to Denavit–Hartenberg parameters $a_5,a_4,d_6$, respectively.
The tool length defines $d_{10}=L_{\mathrm{eff}}$, while the twist
angles satisfy $\alpha_3=\alpha_1$ and $\alpha_6=\beta$.
These analytic relations fully describe how the geometric design variables map into the DH parameters listed in Tab.~\ref{tab:DenavitHartenberg}.

The Denavit–Hartenberg parameters of the $n^\text{th}$ transformation are summarized in Tab.~\ref{tab:DenavitHartenberg}, where $\theta_n^*$ represents the active and passive joint angles $[\phi_{1i}, \phi_{2i}, \phi_{3i}]^T$ for the corresponding leg (see Fig.~\ref{fig:ModifiedCSPM}a). It is evident that $joint~2$ is described in terms of $\phi_{1i}$, $joint~3$ is defined by both ($\phi_{1i}$, $\phi_{2i}$), and the moving platform center $c_p$ as well as the tool contact point $tool$ are determined by the complete set of angles $[\phi_{1i}, \phi_{2i}, \phi_{3i}]^T$.

\begin{table}[t]
\caption{Denavit–Hartenberg table.}
\label{tab:DenavitHartenberg}
\centering
\footnotesize 
\begin{tabular*}{\linewidth}{@{\extracolsep{\fill}}lccccc@{\extracolsep{\fill}}}
\hline\hline
Frame    & Associated Joint     & $\theta$       & $d$       & $a$     & $\alpha$ \\ \hline
$B$      & base      & $0$            & $d_1$     & $0$     & $0$ \\
$j_1$    & joint 1   & $\theta_2^*$   & $d_2$     & $a_2$   & $0$ \\
$j_2$    & joint 2   & $\pi/2$        & $0$       & $0$     & $-\alpha_3$ \\
$-$      &           & $\theta_4^*$   & $0$       & $a_4$   & $\alpha_3$ \\
$-$      &           & $\pi/2$        & $0$       & $a_5$   & $0$ \\
$-$      &           & $\theta_6$     & $d_6$     & $0$     & $\alpha_6$ \\
$j_3$    & joint 3   & $\theta_7^*$   & $0$       & $0$     & $0$ \\
$-$      &           & $0$            & $0$       & $0$     & -$\alpha_6$ \\
$c_p$    & moving platform center    & $\pi/2$        & $0$       & $-a_9$  & $0$ \\
$tool$   & tool tip  & $\theta_{10}$  & $d_{10}$  & $0$     & $0$ \\
\hline\hline
\end{tabular*}

\vspace{0.5em}
\begingroup
\setlength{\baselineskip}{0.9\baselineskip}
\footnotesize
\begin{minipage}{\linewidth}
\justifying
\textit{Note:} $d_1$ is the leg-dependent base offset. $\theta_{n}^{*}$ denotes the active and passive joint angles for the corresponding leg. 
Henceforth, $\theta_{2}^{*}$, $\theta_{4}^{*}$, and $\theta_{7}^{*}$ of each leg are represented by $\phi_{1i}, \phi_{2i}, \phi_{3i}$, where $i = 1, 2, 3$, and the remaining parameters are structural constants that are listed in Table~\ref{tab:MechDesign}. In addition, {$B$}, {$j_1$},{$j_2$}, {$j_3$}, {$c_p$}, and $tool$ are illustrated in Fig.~\ref{fig:ModifiedCSPM}b.
\end{minipage}
\endgroup
\end{table}

For the CDC-SPM mechanism, the \textbf{forward kinematics} defines how the joint variables of each leg determine the pose of the tool frame. Each leg consists of a sequence of five frames: the base frame, the two passive joint frames, the platform-center frame, and finally the tool frame. Since all frame assignments and DH tables have already been introduced, this section simply combines those transformations in the order they appear along each kinematic chain \cite{Craig2022IntroductionControl}, \cite{Spong2020RobotControl}. 

Given the ordered frames $j_0, j_1, j_2, j_3, j_4, j_5$, the complete transformation from frame $a$ to frame $b$ for leg $i$ is written compactly as Eq. \eqref{eq:DirectMapping}:

{\small
\begin{equation}
\begin{gathered}
{}^{b}_{a}T_i 
= \left( \prod_{k=1}^{m} {}^{j_{k-1}}_{j_k}T_i \right),
\qquad \forall ~i\in \{1,2,3\}\\[-2pt]
\text{where based on Fig.~\ref{fig:ModifiedCSPM}b, } j_0 = B,\; j_4 = c_p,\; and\; j_5 = tool.
\end{gathered}
\label{eq:DirectMapping}
\end{equation}
}

Each transformation $ {}^{j_{k-1}}_{j_k}T_i $ in Eq. \eqref{eq:DirectMapping} corresponds directly to a row or derived rows in the DH table \cite{Craig2022IntroductionControl}. The resulting direct transformation is used later in the inverse kinematics and workspace evaluation.

When quantities defined at the tool frame must be mapped back to the base frame or earlier in the leg, \textbf{inverse kinematics} is used.

{\small
\begin{equation}
_{\text{B}}^{\text{tool}} T_{i} = \left(_{\text{tool}}^{\text{B}} T_{i}\right)^{-1}, \quad \forall ~i\in \{1,2,3\}.
\label{eq:Base2Tool}
\end{equation}
}

This is also required in the evaluation of joint coordinates during the inverse kinematics procedure.

\subsubsection{End-Effector Representation}
The representation of the end-effector's tip is essential for solving inverse kinematics, which will be discussed later. Although there are multiple ways to describe this orientation, this study focuses on two methods: quaternions and Yaw-Pitch-Roll (YPR) angles. The inverse kinematics calculations are performed using unit quaternions due to their mathematical efficiency and avoidance of singularities. While quaternions provide precise and robust orientation descriptions, YPR angles are preferred for reporting since they are more intuitive and easier to visualize as sequential rotations.

The \textbf{quaternion} rotation matrix required for transforming the tool orientation into the base frame is \cite{Siciliano2009Robotics}:

{\small
\begin{equation}
R(v) =
\resizebox{0.77\linewidth}{!}{%
$\displaystyle
\begin{bmatrix}
e_0^2 + e_1^2 - e_2^2 - e_3^2  & 2(e_1 e_2 - e_0 e_3)            & 2(e_1 e_3 + e_0 e_2) \\
2(e_1 e_2 + e_0 e_3)           & e_0^2 - e_1^2 + e_2^2 - e_3^2   & 2(e_2 e_3 - e_0 e_1) \\
2(e_1 e_3 - e_0 e_2)           & 2(e_2 e_3 + e_0 e_1)            & e_0^2 - e_1^2 - e_2^2 + e_3^2
\end{bmatrix}
$%
}
\label{eq:Quat_RotationMatrix}
\end{equation}
}

In this study, the rotation matrix $R(v)$ describes the orientation of the end-effector relative to the base frame. From now on, this rotation matrix will be denoted as $\prescript{B}{\text{tool}}{R}_{\text{Quat}}$.

Also, in order to express the tool orientation in an interpretable form for workspace figures and experimental validation, \textbf{Yaw-Pitch-Roll angles (YPR)} are used, which follow the Tait-Bryan convention with a ZYX fixed-angle sequence.
It can be shown that the used ZYX Tait-Bryan convention has singularities in the case where the pitch angle is equal to $\pm \frac{\pi}{2}~\unit{\radian}$. But due to the mechanical restrictions of the manipulator, this angle is limited and cannot reach $\pm \frac{\pi}{2}~\unit{\radian}$; therefore, the singularity never occurs.

\subsection{Optimal Parametric Design for Task-Oriented Medical Applications}

The design of a medical robotic structure must satisfy two tightly coupled domains: tool geometry and task requirements. The objective is to determine structural parameters that ensure full task coverage, collision avoidance, and compact integration.

\subsubsection{Tool Geometry and Mapping Spaces}

The orientation of the tool is described using spherical angles $(\theta, \phi, \psi)$, where $\theta$ represents the tilt angle from the vertical axis, and $\phi$, $\psi$ define rotations about the tool axis. The task defines a bounded tilt range:
{\small
\begin{equation}
\theta \in [-\theta_{\min}, \theta_{\min}]
\label{eq:Eq1_ParametricDesign}
\end{equation}
}

This task-specific angular range, as illustrated for ultrasound scanning in Fig.~\ref{fig:UltrasoundProbeWorkspace}, defines a conical workspace that the mechanism must achieve. These angles can be interpreted as $\alpha_{1}$ (see Fig.~\ref{fig:ModifiedCSPM}a).

Let $L_{\text{tool}}$ denote the total length of the medical tool, and $L_{\text{eff}}$ represent the effective portion of $L_{\text{tool}}$ located between the moving platform and $CoR$. Define two circles with radii $R_1$ and $R_2$, on whose circumferences $joints~2$ and $joints~3$ are located, respectively (Fig.~\ref{fig:ModifiedCSPM}c). To ensure that the tool can move freely without colliding with any links or joints, it is assumed that $L_{\text{tool}}$ is positioned above the height corresponding to the circle of radius $R_1$. Under this assumption, the inner and outer radii of the workspace can be expressed as follows:

{\small
\begin{equation}
R_1 = L_{tool} \tan(\theta_{\min}), \quad
R_2 = L_{eff} \tan(\beta)
\label{eq:Eq2_ParametricDesign}
\end{equation}
}

To avoid interference, the calculated radius $R_2$ shall be greater than the tool cross-sectional radius at the $L_{eff}$ height, denoted by $r_{eff}$. If $R_2$ is smaller than $r_{eff}$, a larger radius must be selected. It should be noted that the size of $R_2$ has a direct relationship with the angle $\beta$. Specifically, a larger $R_2$ results in a larger $\beta$, which can limit the maximum tilt angle of the moving platform, ($\theta_{\max}$). Therefore, a design trade-off exists between these parameters. To achieve an optimal design, it is generally preferable to select $R_2$ as close as possible to the radius of the tool ($r_{eff}$). This choice helps balance the competing effects while avoiding unnecessary restrictions on the platform's tilt capability.

Additionally, as illustrated in Fig.~\ref{fig:ModifiedCSPM}c, another important parameter denoted by $\xi$ represents the rotation angle between the base and the moving platform. This parameter significantly influences the maximum achievable tilt angle. In general, increasing $\xi$ leads to a larger attainable $\theta_{\max}$. According to \cite{Li2025Closed-formAxes}, setting $\xi = \frac{\pi}{2}~\unit{\radian}$ simplifies the kinematic constraints considerably. In this configuration, terms such as $\sin(\xi)$ reach their maximum value, while $\cos(\xi)$ becomes zero. As a result, the coefficient expressions are significantly simplified. Notably, the forward kinematics problem can be reduced from solving a quartic equation to a quadratic one, enabling more efficient and robust closed-form solutions. Therefore, $\xi = \frac{\pi}{2}~\unit{\radian}$ is often considered an optimal choice, although values slightly above or below this may also be considered depending on specific design requirements.

\subsubsection{Collision Avoidance Constraints}
Following the geometric design of $R_1$ and $R_2$, additional safety margins are required to prevent collisions with the joints and surrounding components.

{\small
\begin{equation}
R_1' = R_1 + r, \quad
R_2' = R_2 + r
\label{eq:Eq3_ParametricDesign}
\end{equation}
}

The parameter $r$ represents the radius of the joints, which is assumed to be identical and sufficiently small for all joints. In addition, to prevent collision between the tool and the base, the parameter $d_{2}$ is defined as the height of a circle with radius $R_{1}$ measured from the base.

After determining all required parameters, the vertical position of the center of rotation $z_{CoR}$ can be computed as:

{\small
\begin{equation}
z_{CoR} = d_{1} + d_{2} + d_{6} + L_{eff} = d_{1} + d_{2} + L_{tool}
\label{eq:Eq4_ParametricDesign}
\end{equation}
}

$d_{1}$, $d_{2}$, $d_{6}$, and $CoR$ is shown in Fig.~\ref{fig:ModifiedCSPM}c. These parameters correspond to the same quantities whose relationships were previously derived in earlier sections. In particular, $d_{2}$ and $d_{6}$ represent the third components of the vectors $\vec{r}_{j_1j_2}$ and $\vec{r}_{j_2j_3}$ respectively, expressed in terms of $z_{CoR}$ (Fig.~\ref{fig:ModifiedCSPM}b).

The exact position of $joint~3$ relative to $joint~2$ can be determined once the rotation of the moving platform about the x-axis with respect to the base is specified; this parameter is denoted by $\xi$ in Fig.~\ref{fig:ModifiedCSPM}c. Based on this geometric configuration, the coupling between the orientations of the base and the moving platform can be established.

\subsubsection{Orientation Coupling}
Once the orientation parameters are defined, the angle $\alpha_2$ can be determined. Based on the spherical law of cosines, the relationship between the base and moving platform orientations is expressed in Eq.~\ref{eq:Eq5_ParametricDesign}. This relation is derived from the spherical triangle defined by the joint axes.

{\small
\begin{equation}
\cos (\alpha_2) =
\cos (\alpha_1) \cos(\beta) +
\sin (\alpha_1) \sin (\beta) \cos (\xi)
\label{eq:Eq5_ParametricDesign}
\end{equation}
}

Accordingly, the angle $\alpha_2$ can be obtained as:

{\small
\begin{equation}
\alpha_2 =
\cos^{-1}
\left(
\cos (\alpha_1) \cos (\beta) +
\sin (\alpha_1) \sin (\beta) \cos (\xi)
\right)
\label{eq:Eq6_ParametricDesign}
\end{equation}
}

This equation indicates that the orientation $\alpha_2$ is a function of the reference angle $\alpha_1$, the constant geometric tilt $\beta$, and the azimuthal rotation $\xi$ (see Fig.\ref{fig:ModifiedCSPM}a and Fig.\ref{fig:ModifiedCSPM}c). This relation highlights the coupling between the elevation and rotational degrees of freedom of the mechanism. The obtained equations are used in the following section to analyze the kinematic behavior of the system and to develop the CAD models of the application tools listed in Tab.~\ref{tab:MechDesign}.

\subsection{Kinematics Models Implementation}
Using MATLAB (Mathworks, US), the CDC-SPM’s kinematics were analyzed via homogeneous transformations. Forward kinematics is performed by transformation through the tool frame, whereas inverse kinematics is based on transformation through the base frame. 

\subsubsection{Forward Kinematics}
Forward kinematics is crucial for determining the precise orientation of the tool platform. In this process, the motor angles $\phi_{1i}, i=1,2,3$ are known, and the goal is to compute the orientation of the tool in the quaternion format. $q = [e_0, e_1, e_2, e_3]$. To resolve this system, four equations must be established. Considering the manipulator’s closed kinematics chain, Eq.~\ref{eq:Eq1_ForwardKinematics} is formulated, where $v_i$ represents the $z$-direction of the first passive joint ($joint~2$) and $w_i$ represents the $z$-direction of the second passive joint ($joint~3$). 

{\small
\begin{equation}
v_i \cdot w_i = \cos(\alpha_2), \quad \forall ~i\in \{1,2,3\}.
\label{eq:Eq1_ForwardKinematics}
\end{equation}
}

If the vector $v_i$ is expressed with respect to the base frame and the vector $w_i$ with respect to the tool frame, the relation can be rewritten as:
{\small
\begin{equation}
v_i\big|_{\text{via base}} \cdot w_i\big|_{\text{via tool}} = \cos(\alpha_2), \quad \forall ~i\in \{1,2,3\}.
\label{eq:Eq2_ForwardKinematics}
\end{equation}
}

where $v_i\big|_{\text{via base}}$ is obtained from the third column of its transformation matrix, transforming the base frame to $joint~2$, expressed purely in terms of the motor angle $\phi_{1i}$. Also, $w_i\big|_{\text{via tool}}$ is extracted from the third column of the transformation matrix, transforming the tool frame to $Joint~3$, expressed in terms of the quaternion elements. $\alpha_2$ is a known structural parameter that defines the curvature of the second link. By substituting $v_i(\phi_{1i})$ and $w_i(e_0, e_1, e_2, e_3)$, Eq.~\ref{eq:Eq2_ForwardKinematics} can be rewritten as a function:

{\small
\begin{equation}
f_i \left(e_0, e_1, e_2, e_3, \phi_{1i}\right) = 0, \quad \forall ~i\in \{1,2,3\}.
\label{eq:Eq3_ForwardKinematics}
\end{equation}
}

These transformations generate three equations when specific values for $\phi_{1i}$ are inserted in the function. The fourth equation is derived from the unit quaternion property:

{\small
\begin{equation}
e_0^2 + e_1^2 + e_2^2 + e_3^2 = 1.
\label{eq:Eq4_ForwardKinematics}
\end{equation}
}

This constraint ensures that the quaternion remains valid as a rotation representation. By solving this set of equations, the quaternion components can be determined for any given set of motor angles $\phi_{1i}$.

\subsubsection{Inverse Kinematics}
The inverse kinematics problem consists of uniquely determining the input angles $\phi_{1i}$ required to achieve a specified orientation of the end-effector, represented in quaternion format $q = [e_0, e_1, e_2, e_3]$. The angles of the intermediate passive joints ($joint~2$ and $joint~3$) are not strictly necessary for solving the inverse kinematics, but they can be calculated if needed.

\vspace{0.5em}
\textbf{Calculation of $\phi_{1i}$}:
To calculate the angles of the active joints $\phi_{1i}$, Eq.~\ref{eq:Eq1_ForwardKinematics} is still applied but, this time, the rotation matrix extracted from the inverse transformation of the tip to $joint~3$ is now known since the desired orientation in quaternion form $q = [e_0, e_1, e_2, e_3]$ is provided. This allows the direct computation of $w_i\big|_{\text{via tool}}$, which is extracted from the third column of the transformation matrix using the known quaternion components. This time, $v_i\big|_{\text{via base}}$ will be expressed purely in terms of the motor angle $\phi_{1i}$. Substituting $v_i(\phi_{1i})$ and the precomputed $w_i$ into Eq.~\ref{eq:Eq1_ForwardKinematics} simplifies it into a new function:

{\small
\begin{equation}
f_i \left(\phi_{11}, \phi_{12}, \phi_{13} \right) = 0, \quad \forall ~i\in \{1,2,3\}.
\label{eq:Eq2phi1_Inverseinematics}
\end{equation}
}

Equation~\ref{eq:Eq2phi1_Inverseinematics} results in a system of three equations with three unknowns $[\phi_{11}, \phi_{12}, \phi_{13}]^T$. By solving this system, the necessary joint angles $\phi_{1i}$ can be determined for any given quaternion orientation. Additionally, the angles of intermediate passive joints can be determined in the same manner.

\vspace{0.5em}
\textbf{Calculation of $\phi_{2i}$}:
Now, the rotation matrix obtained from direct transformation of the $base$ to $joint~3$ is known, as the angles of the active joints $\phi_{1i}$ have been calculated in the previous step. This enables the direct extraction of $w_i$, which corresponds to the third column of the transformation matrix.

Considering that the rotation matrix representing each joint frame must be unique, the extracted $w_i$ must be equal to the $z$-axis direction obtained from an inverse transformation (from the $tool$ to $joint~3$). In other words, the following equation must hold:

{\small
\begin{equation}
w_i\big|_{\text{via tool}} - w_i\big|_{\text{via base}} = 0, \quad \forall ~i\in \{1,2,3\}.
\label{eq:Eq1phi_Inverseinematics}
\end{equation}
}

\noindent Here, $w_i\big|_{\text{via base}}$ is known, while $w_i\big|_{\text{via tool}}$ is computed using Eq.~\ref{eq:Eq2phi_Inverseinematics}:

{\small
\begin{equation}
w_i\big|_{\text{via tool}} = \prescript{B}{\text{tool}}{R}_{\text{Quat}} \cdot \prescript{\text{tool}}{j_3}{R}_i, \quad \forall ~i\in \{1,2,3\}.
\label{eq:Eq2phi_Inverseinematics}
\end{equation}
}

Substituting Eq.~\ref{eq:Eq2phi_Inverseinematics} into Eq.~\ref{eq:Eq1phi_Inverseinematics} yields a system of three equations with three unknowns $[\phi_{21}, \phi_{22}, \phi_{23}]^T$. Solving this system provides the required joint angles for the first passive joints $\phi_{2i}$.

\vspace{0.5em}
\textbf{Calculation of $\phi_{3i}$}:
Again, considering that the rotation matrix representing each joint frame must be unique, the $z$-axis direction extracted from the end-effector transformation matrix, expressed using quaternions and denoted as $n_i\big|_{\text{via tool}}$, must match the $z$-axis direction obtained from the direct transformation (from the $base$ to the $tool$), denoted as $n_i\big|_{\text{via base}}$. This condition leads to the determination of the required joint angles for the second passive joints $\phi_{3i}$, as expressed by the following equation:

{\small
\begin{equation}
n_i\big|_{\text{via tool}} - n_i\big|_{\text{via base}} = 0, \quad \forall ~i\in \{1,2,3\}.
\label{eq:phi3_Inverseinematics}
\end{equation}
}

In which $n_i\big|_{\text{via tool}}$ corresponds to the third column of $\prescript{B}{\text{tool}}{R}_{\text{Quat}}$ and is known since the quaternion components are given. In contrast, $n_i\big|_{\text{via base}}$ represents the third column of the direct transformation from the $base$ to the $tool$ through each leg, and this transformation is a function of the angles $\phi_{1i}, \phi_{2i},$ and $\phi_{3i}$. After substituting the previously computed angles $\phi_{1i}, \phi_{2i}$, the expression for $n_i\big|_{\text{via base}}$ becomes dependent only on $\phi_{3i}$. As a result, Eq.~\ref{eq:phi3_Inverseinematics} forms a system of three equations with three unknowns $[\phi_{31}, \phi_{32}, \phi_{33}]^T$. Solving this system provides the required joint angles for the second passive joints $\phi_{3i}$.

\subsection{Kinematic Performance Analysis}
There are certain positions that the robot finds difficult to reach because they correspond to singularities. At these singular points, the system becomes highly sensitive, meaning that a small change in joint angles can cause a large change in the end-effector’s position or orientation. The \textbf{Jacobian matrix} $J$ is a powerful tool for analysing this behavior throughout the workspace.

For parallel robotic mechanisms, the Jacobian analysis differs from that of serial robots because the kinematic constraints couple the joint motions \cite{Lynch2017MODERNCONTROL}. 

When the loop-closure constraint equation:
{\small
\begin{equation}
    F(\boldsymbol{x}, \boldsymbol{q}) = 0,
    \label{eq:Eq1_JacobianAnalysis}
\end{equation}
}
where $x$ denotes the end-effector pose variables [yaw, pitch, roll], and $q$ represents the actuated joint variables $[\phi_{11}, \phi_{12}, \phi_{13}]^T$, is differentiated, it yields the general constraint equation:
{\small
\begin{equation}
    J_{x}\, \dot{\boldsymbol{x}} + J_{q}\, \dot{\boldsymbol{q}} = 0.
    \label{eq:Eq2_JacobianAnalysis}
\end{equation}
}

which produces two Jacobian matrices \cite{Zhang2010ParallelTools}:
\begin{itemize}
    \item \textbf{$J_{x}$}: relating changes in end-effector pose to the constraints,
    \item \textbf{$J_{q}$}: relating changes in the actuated joint to the constraints.
\end{itemize}

Solving for the end-effector velocity gives the effective Jacobian of the parallel mechanism ($J$):

{\small
\begin{equation}
    \dot{\boldsymbol{x}} = J \, \dot{\boldsymbol{q}}, 
    \qquad 
    J = - J_{x}^{-1} J_{q},
    \label{eq:Eq3_JacobianAnalysis}
\end{equation}
}

This effective Jacobian $J$ plays the same role as in a serial robot, but it already embeds the closed-loop constraints of the parallel structure. It captures the complete constrained kinematic behavior of the mechanism and is used for analysing singularities and manipulability. Singular configurations occur when the Jacobian matrix $J$ becomes rank-deficient, leading to a loss of motion capability or controllability of the mechanism. Rather than explicitly classifying these singularities, their effect can be evaluated through a numerical performance measure.

The manipulability of a configuration refers to how effectively the robot can generate end-effector motion from joint motion \cite{Siciliano2009Robotics}. A well-established measure of manipulability is the condition number of the Jacobian, denoted as $\kappa(J)$\cite{Gosselin1993OnManipulators}.

{\small
\begin{equation}
\kappa(J) = \| J \| \cdot \| J^{-1} \|
\label{eq:Eq4_JacobianAnalysis}
\end{equation}
}

\noindent Here, $\| J \|$ and $\|J^{-1}\|$ represent the maximum stretching effects of the Jacobian and its inverse, respectively. Their product provides a measure of the worst-case sensitivity, yielding the condition number in the range $1 \leq \kappa(J) < \infty$. These norms can be calculated using the following expressions:

{\small
\begin{equation}
\| J \| = \sqrt{\text{tr}(J^T W J)}, \qquad
\| J^{-1} \| = \sqrt{\text{tr}\!\left((J^{-1})^T W J^{-1}\right)} 
\label{eq:Eq5_JacobianAnalysis}
\end{equation}
}

In which, the matrix $W$ is defined as $W = \frac{1}{n}I$, where $n$ is the dimension of the square matrix $J$, and $I$ denotes the $n \times n$ identity matrix. In this study, $n = 3$ corresponds to the dimension of the mechanism. To simplify the numerical computations, the reciprocal of the condition number is defined as:
{\small
\begin{equation}
\xi(J) = \frac{1}{\kappa(J)},
\label{eq:Eq7_JacobianAnalysis}
\end{equation}
}
which implies $0 < \xi(J) < 1$. Thus, $\xi(J) \in (0,1)$ is used to evaluate the conditioning:

\begin{itemize}
    \item If $\xi(J) \approx 1$, the system is well-conditioned, indicating stability and insensitivity to small variations in joint angles.
    \item If $\xi(J) \approx 0$, the system is ill-conditioned, meaning even a tiny change in the input can cause large, unpredictable changes in the output.
\end{itemize}
The insights from the kinematic performance analysis guide the selection of geometric parameters and design choices presented in the following section.

\subsection{Detailed Mechanical Design}
\subsubsection{CAD Design}
The proposed framework, described in the previous sections, is applied to the following four ultrasound probes:

{\small
\begin{itemize}
\item CERBERO 4.0 B-MatrixSound Convex Ultrasound Probe
\item Butterfly iQ Linear Ultrasound Probe
\item C10RC Endovaginal + Convex + Cardio Handheld Wireless Probe
\item BladderScan i10 Ultrasound Scanner with Automated Bladder Volume Measurement
\end{itemize}
}

\noindent to demonstrate how the derived equations enable the design of an optimal CDC-SPM interface by incorporating their geometric and functional requirements.

\begin{figure*}[!t]
\centering
\includegraphics[width=\textwidth]{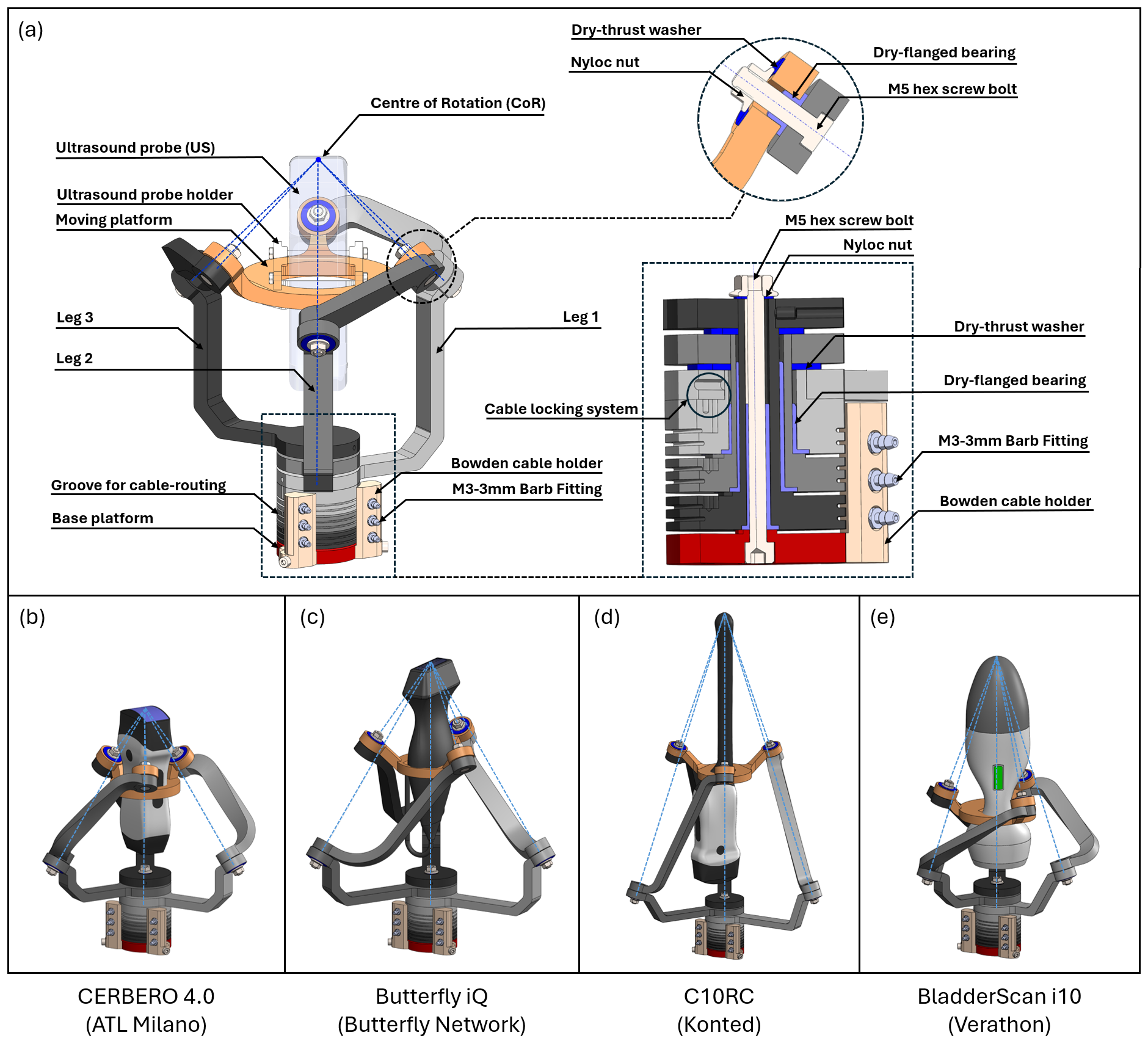}
\caption{CAD model of the proposed CDC-SPM: (a-left) isometric view, (a-right) section view, and designing for some medical ultrasound probes (b) ATL Milano CERBERO 4.0, (c) Butterfly Network iQ, (d) Konted C10RC, (e) Verathon BladderScan i10}
\label{fig:SolidWorksModel}
\end{figure*}

Figure~\ref{fig:SolidWorksModel} shows the 3D model of the CDC-SPM mechanism, designed using SolidWorks (Dassault Systèmes SE, FR) with the parameters listed in Tab.~\ref{tab:MechDesign} for case study ultrasound probes (Fig.~\ref{fig:SolidWorksModel}b--d), as well as a detailed CAD representation of the interface (Fig.~\ref{fig:SolidWorksModel}a). This figure provides a clear visualization of the overall structure and geometry of the mechanism, serving as the foundation for both simulation and physical prototyping. As depicted in Fig.~\ref{fig:SolidWorksModel}a, the rotational axes of all joints meet at a single central point, referred to as the center of rotation. 

\begin{table*}[!t]
\caption{Parameters for mechanical system design.}
\label{tab:MechDesign}
\scriptsize
\centering

\begin{tabular*}{\textwidth}{@{\extracolsep{\fill}}l l p{5.5cm} c c c c c@{}}
\hline\hline
\textbf{Variable} & \textbf{DH Parameter} & \textbf{Description} & \textbf{Unit} 
& \makecell{\textbf{ATL Milano}\\ \textbf{CERBERO 4.0}} 
& \makecell{\textbf{Butterfly Network}\\ \textbf{Butterfly iQ}} 
& \makecell{\textbf{Konted}\\ \textbf{C10RC}} 
& \makecell{\textbf{Verathon}\\ \textbf{BladderScan i10}} 
\rule{0pt}{8pt}\\
\hline

$\alpha_{1}$         & $\alpha_{3}$                    & Angle between z-axis of joint 1 $\&$ joint 2 (link1 curvature)      & rad   & $0.61$     & $0.61$     & $0.35$      & $0.35$\\
$\alpha_{2}$         & $*$                             & Angle between z-axis of joint 2 $\&$ joint 3 (link2 curvature)      & rad   & $0.59$  & $0.84$  & $0.34$   & $0.49$\\
$\beta$              & $\alpha_{6}$                    & Angle between z-axis of joint 3 $\&$ $\vec{n}$ of moving platform  & rad   & $0.62$  & $0.62$  & $0.35$   & $0.35$\\
$\xi$                & $-\theta_{10}$                  & Angle between x-axis of the moving platform $\&$ the base         & rad   & $\frac{\pi}{3}$     & $\frac{\pi}{2}$     & $\frac{\pi}{3}$      & $\frac{\pi}{2}$\\
$L_{\mathrm{tool}}$  & $*$                             & Total length of medical tool                                      & mm    & $121$     & $152$    & $310$      & $194$\\
$L_{\mathrm{eff}}$   & $d_{10}$                        & Effective length of medical tool attachable to this mechanism     & mm    & $50$     & $72$   & $170$    & $130.5$\\
$r_{\mathrm{eff}}$   & $*$                             & Effective radius of medical tool attachable to this mechanism     & mm    & $27$   & $30$   & $8.25$   & $30$\\
$R_{1}$              & $a_{2}$                         & Radius of the circle on which joints 2 are lied                    & mm    & $85$     & $107$     & $113$     & $71$\\
$R_{2}$              & $a_{9}$                         & Radius of the circle on which joints 3 are lied                    & mm    & $36$     & $51$      & $62$      & $48$\\
$d_{2}$              & $d_{2}$                         & Height of joints 2 with respect to the base                        & mm    & $60$
  & $49.5$ & $40$     & $40$\\
$z_{CoR}$            & $d_{1}+d_{2}$+$d_{6}$+$d_{10}$  & Center of rotation (tip of medical tool)                          & mm    & $184.5$ & $205$  & $353.5$  & $237.5$\\
$offset$             & $d_{1}$ (leg 1)                  & Offset of the first active joint from the base                    & mm    & $3.5$    & $3.5$  & $3.5$    & $3.5$\\

\hline\hline
\end{tabular*}

\vspace{4pt}
\justifying
\scriptsize
$*$ \textit{Note:} $\alpha_{2}$ and $z_{CoR}$ do not appear as single DH parameters. 
Instead, they define the spatial geometry of link~2 and link~3, which produces the intermediate DH parameters 
$a_{2}$, $a_{4}$, $a_{5}$ and $d_{2}$, $d_{6}$, $d_{10}$ (see Table~\ref{tab:DenavitHartenberg}). Assuming a 8.5 mm gap between coaxial active joints, the offsets for leg 2 and leg 3 are 12 mm and 20.5 mm..  
\normalsize
\end{table*}

Each of the three legs moves individually via a coaxial pulley system in a lightweight, cable-driven design. This coaxial configuration offers a balanced and minimal footprint for the mechanism with a single cable management system. Dry bearings have been chosen to minimise the weight and size of the mechanisms. Additionally, this self-lubricating joint design reduces maintenance and is suitable for repeated actuations. The current prototype is 3D printed, but the final mechanism is going to be in aluminium, keeping the moving mass around 550 grams.

Each pulley design includes a built-in cable locking system that uses the pulley mounting screws to secure the cable. The pulley has been dimensioned to ensure that the polymeric rope routing to the Bowden cable is reliable, maintaining the tension and alignment across its entire range of movement. PTFE tubes are then used as a Bowden cable to route the polymeric rope to the motors that are placed on a separate support. This configuration allows for reducing the moving mass when the proposed haptic interface is mounted on the robotic arm. Thus, reducing the load at the end-effector for the robot arm will increase the responsiveness and efficiency of the final robot.

\begin{figure}[!htbp]
\centering
\includegraphics[width=\columnwidth]{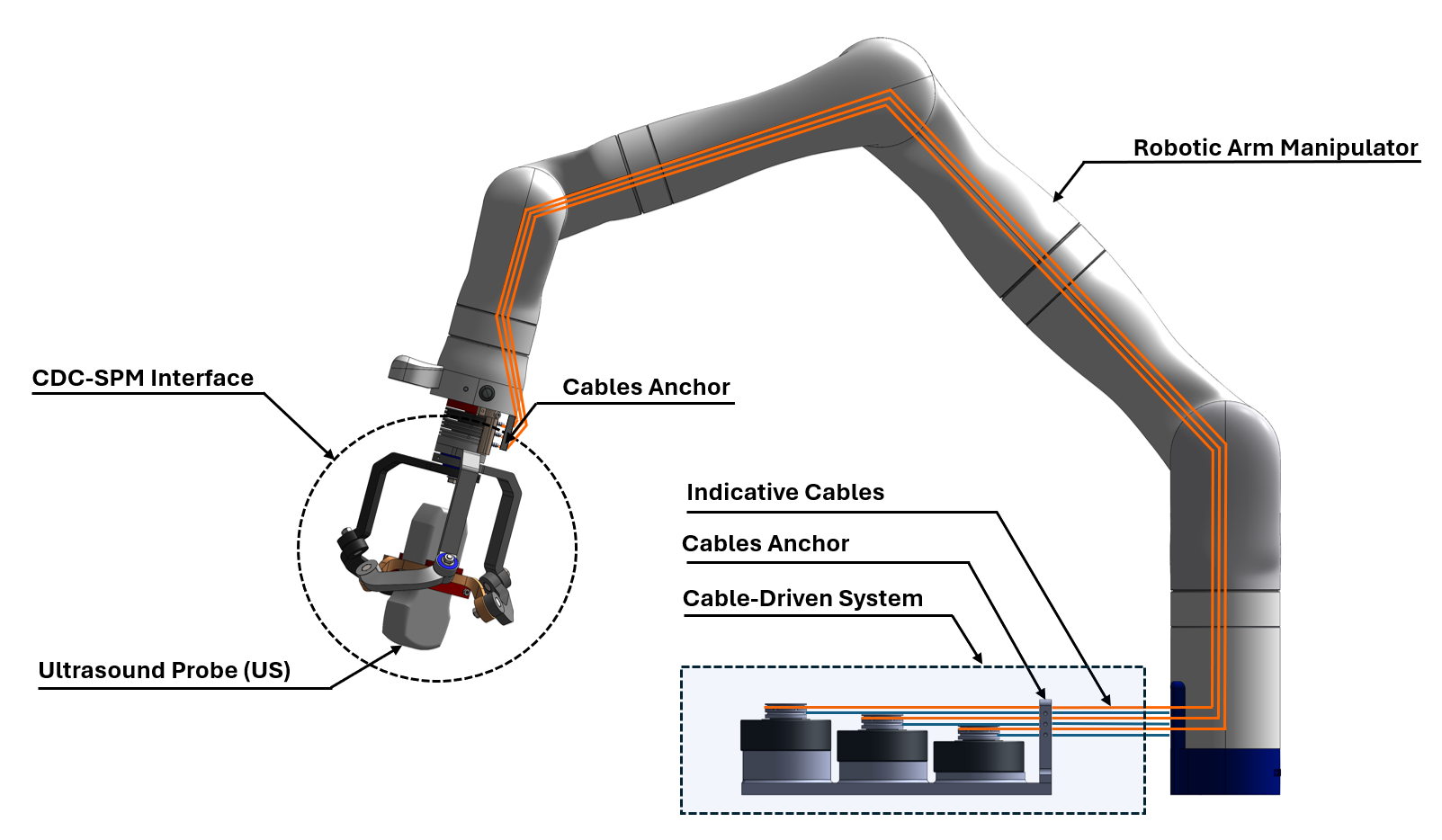}
\caption{Detailed integration of the actuation system and end-effector attachment.}
\label{fig:DetailedDesign}
\end{figure}

The attachment of the proposed interface to a sample robotic manipulator via the end-effector is illustrated in Fig.~\ref{fig:DetailedDesign}, highlighting its compatibility as a modular component within a larger robotic system. The detailed integration of the actuation system and end-effector attachment is also shown. Motors are secured to the base structure and arranged to enable a cable-driven transmission, allowing their mass to remain isolated from the moving components and thereby reducing the reflected inertia. The ultrasound probe is mounted to the CDC-SPM interface using a dedicated holder that ensures stable fixation while preserving the required rotational motion about the defined center of rotation.

\subsubsection{Finite Element Analysis}
To assess the structural integrity of the CDC-SPM, Finite Element Analysis (FEA) was performed using industry-standard methods in ANSYS Workbench (ANSYS, Inc., US). The CAD model was imported and assigned relevant material properties corresponding to Aluminum 6061-T6. The system was then simulated under boundary conditions, including a fixed robot frame and roller-type constraints at the joints. To represent the mechanical demands of our case study (ultrasound scanning), a $50\,\mathrm{N}$ force was applied at the center of rotation (CoR) of the end-effector, alongside gravity.

\subsubsection{Test Bench Design}
\begin{figure}[!htbp]
\centering
\includegraphics[width=\columnwidth]{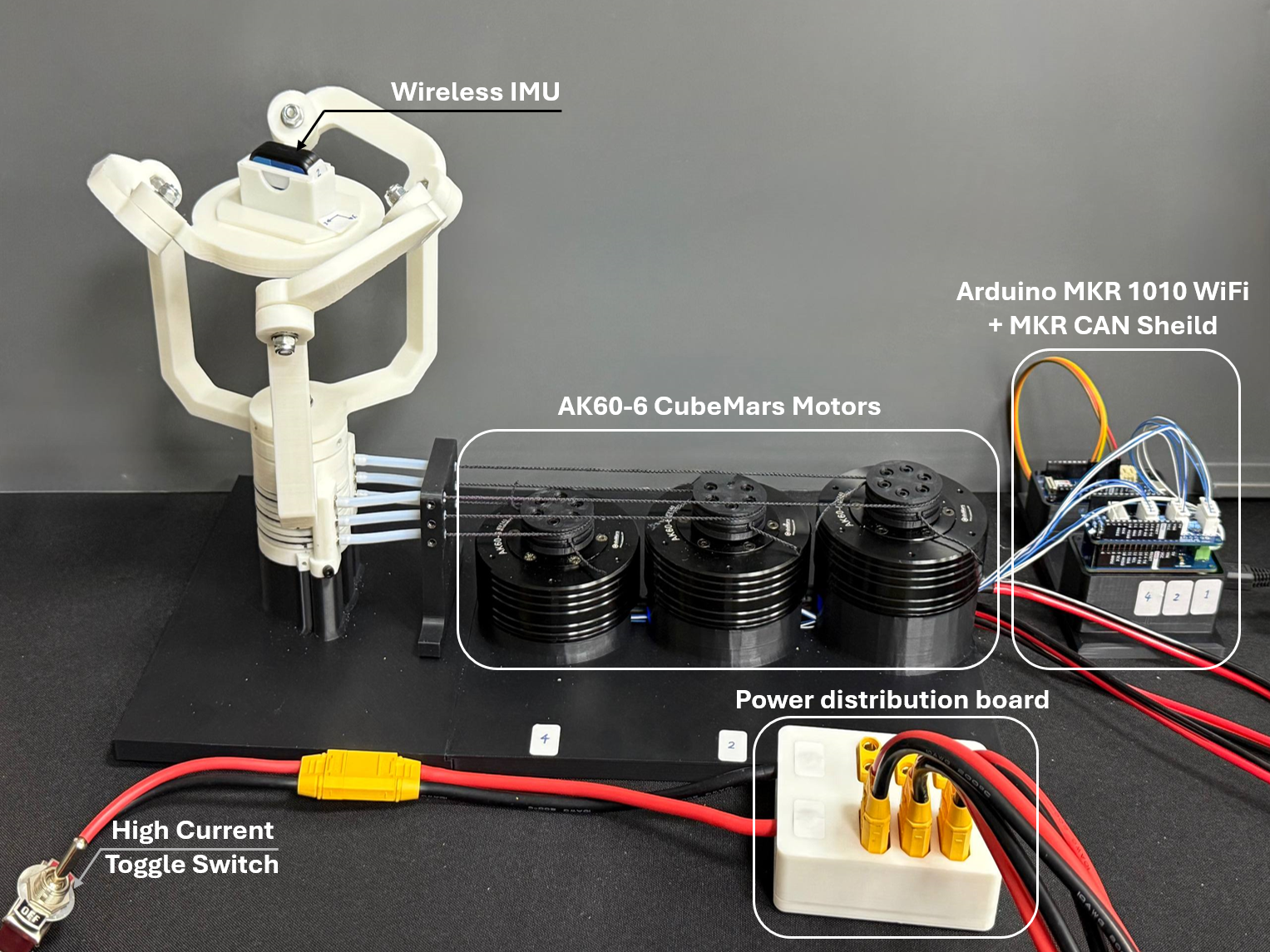}
\caption{The first prototype fabricated in PLA by 3D printing.}
\label{fig:Prototype}
\end{figure}

The first prototype fabricated in PLA by 3D printing is shown in Fig.~\ref{fig:Prototype}. While this version serves as a proof-of-concept platform, the final design will be manufactured from lightweight Aluminum alloy to improve stiffness, structural durability, and load capacity. 

The system is actuated by three AK60-6 CubeMars motors, controlled over a CAN bus network using an Arduino MKR 1010 WiFi paired with an MKR CAN Shield. A WT9011DCL 9-axis wireless IMU module with a stated angular accuracy of $0.2~\unit{\deg}$ was mounted on the moving platform to measure its orientation in real time. Both the IMU data acquisition and the motor position commands were executed at $200~\unit{\hertz}$. To characterize the mechanism’s behavior, forward and inverse kinematic models were developed in MATLAB. Since the mechanism features complex geometry and tight joint spacing, the forward kinematics model also includes an additional three-step procedure to identify self-collisions and unfeasible poses:
\begin{enumerate}
    \item Configurations where the absolute difference between any two joints is either too small or too large $\left(\frac{\pi}{10}~ < |\Delta \phi_{1i}| < \pi \right)$ are marked as potential self-collision cases and immediately rejected. The threshold of $\frac{\pi}{10}~\unit{\radian}$ is chosen as a safety margin to avoid physical interference between the legs of the mechanism.
    \item For the configurations that pass the collision check, forward kinematics is computed to obtain the end-effector orientation in terms of yaw-pitch-roll (YPR). If any of these values are invalid or undefined, the configuration is considered non-feasible and discarded.
    \item The Jacobian matrix and its condition number are calculated. Configurations with reciprocal condition number below a predefined threshold $\xi_{\min}(J) = 0.2$ are classified as near-singular and are excluded from the feasible set.
\end{enumerate}
The remaining configurations, which satisfy both collision and singularity criteria, form the final feasible configuration space.

To validate if the proposed mechanism has a suitable range of motion for ultrasound applications, we experimentally compare the theoretical mechanism's range of motion with the prototype's range of motion. This comparison is needed due to the complex mechanism geometry that does not allow an easy evaluation of the range of motion limitations imposed by the self-collisions. The inverse kinematics solver is initially used to determine the actuated joint angles corresponding to desired roll, pitch, and yaw values for the ultrasound task. As described by Fig.~\ref{fig:UltrasoundProbeWorkspace}, they are $\pm 35~\unit{\deg}$ ($\pm 0.611~\unit{\radian}$) for the roll, pitch, and $\pm 180~\unit{\deg}$ ($\pm \pi~\unit{\radian}$) for the yaw. These values were used as input for the direct kinematic model and the test-bench motors, allowing for comparison of the theoretical and real workspace.  

\section{Results}
\begin{figure*}[htbp]
\centering
\includegraphics[width=1\textwidth]{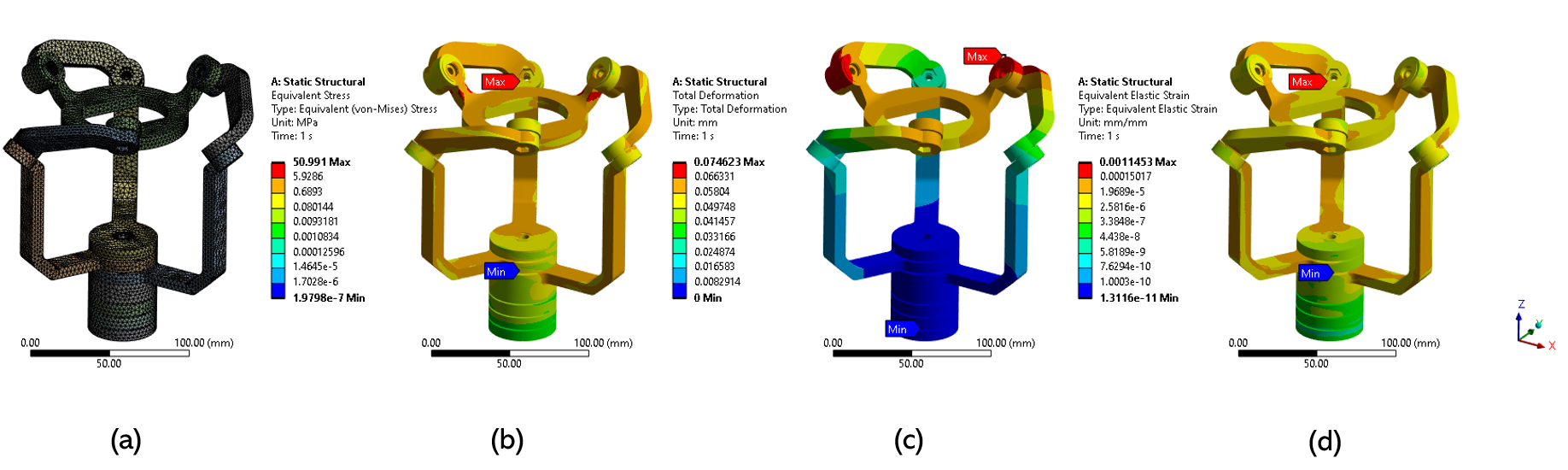}
\caption{Finite element analysis: (a) Model meshing, (b) Total deformation, (c) Von Mises stress, (d) Elastic strain.}
\label{fig:FEMAnalysis}
\end{figure*}

To validate the structural performance of the proposed design, the FEA model is discretized with a refined $3\,\unit{mm}$ mesh, creating 270,216 nodes and 157,895 elements, as shown in Fig.~\ref{fig:FEMAnalysis}a. Maximum von Mises stress appeared at the middle joints but stayed below $51\,\mathrm{MPa}$ (Fig.~\ref{fig:FEMAnalysis}b), confirming that Aluminum is a safe material choice with a minimum safety factor of 5.5. Total deformation remained well under operational limits, with a maximum of $0.075\,\mathrm{mm}$ (Fig.~\ref{fig:FEMAnalysis}c), while static strain was negligible (Fig.~\ref{fig:FEMAnalysis}d).

\begin{figure}[htbp]
\centering
\includegraphics[width=\columnwidth]{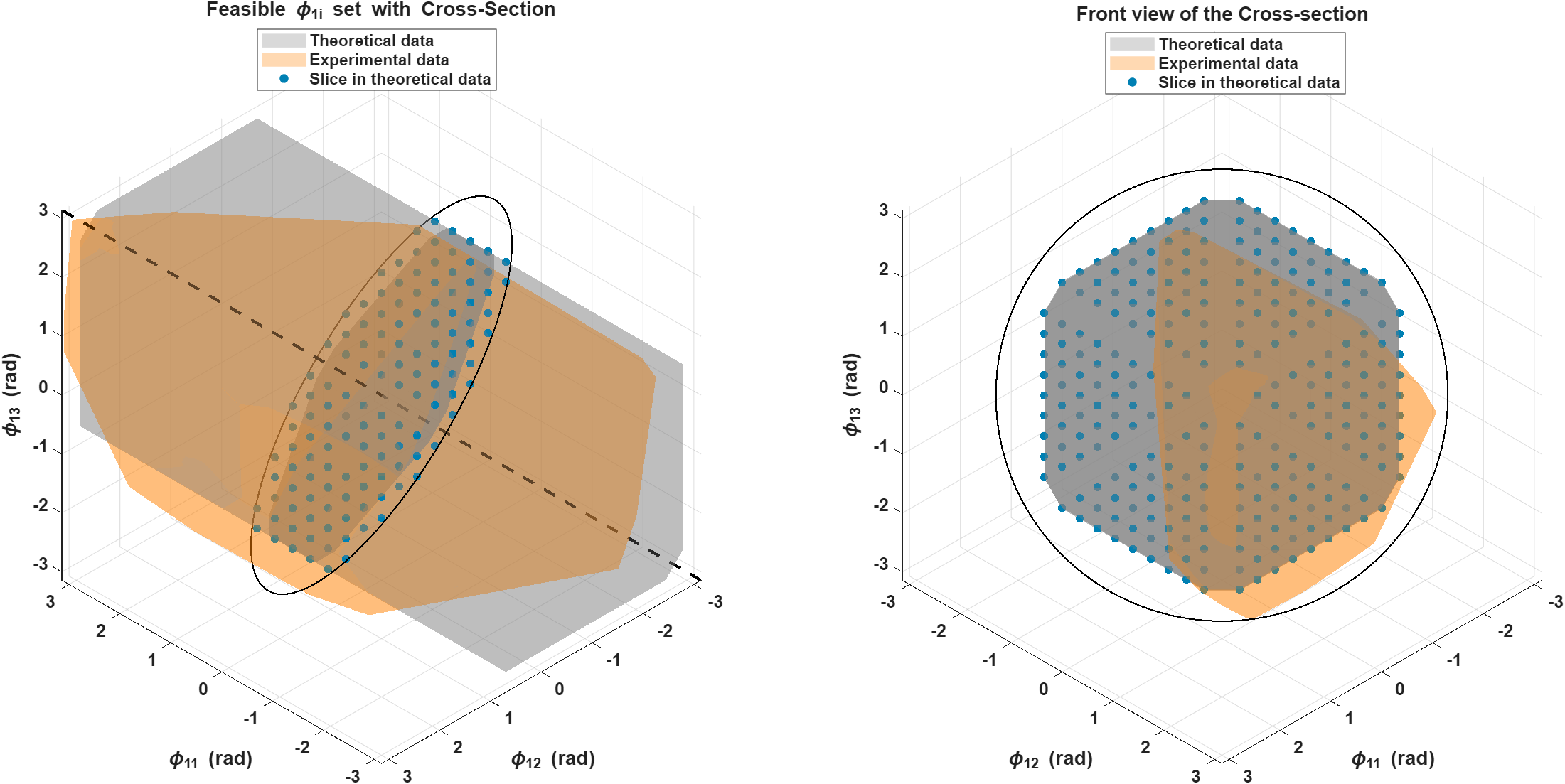}
\caption{Numerically computed configuration space of CDC-SPM (gray) compared with experimentally measured (orange).}
\label{fig:ConfigurationSpace}
\end{figure}

The feasible configuration space of the structure is evaluated by sampling the input joint vector $\phi_{1i}$ where each element is in the interval $[-\pi, \pi]~\unit{\radian}$ with a discretization step of $\frac{\pi}{18}~\unit{\radian}$, resulting in 50,653 test configurations. Each configuration was validated to exclude singular and near-singular states, as well as link collisions. The resulting feasible set forms a constrained subset of a cubic domain shown by the gray convex hull in Fig.~\ref{fig:ConfigurationSpace}. A key observation is that the feasible configuration space exhibits a constant cross-sectional profile along the diagonal direction connecting $\phi_{1i} = [-\pi, -\pi, -\pi]^T$ and $\phi_{1i} = [\pi, \pi, \pi]^T$ (Fig.~\ref{fig:ConfigurationSpace}). This direction corresponds to synchronized actuation of all joints and represents pure rotational motion of the mobile platform about its vertical axis. The invariance of the cross-section along this direction indicates that the mechanism supports continuous (theoretically unbounded) rotation about this axis. Furthermore, the feasible configurations are bounded such that deviations of individual joint values from this diagonal trajectory do not exceed approximately $0.77\pi~\unit{\radian}$. This implies that relative differences between actuator inputs are limited, as larger deviations lead to link interference. Consequently, the configuration space can be interpreted as a tube-like region extending along the diagonal direction.

The Cartesian orientation workspace is computed using a discretization of the tool orientation vector $\mathrm{YPR}$, where yaw $\in [-\pi, \pi]$, pitch $\in [-\frac{\pi}{3}, \frac{\pi}{3}]$, and roll $\in [-\frac{\pi}{3}, \frac{\pi}{3}]$, using a step size of $\frac{\pi}{36}~\unit{\radian}$  for roll and pitch and $\frac{\pi}{18}~\unit{\radian}$ for yaw, resulting in 50,653 test orientations. At each orientation, inverse kinematics solutions containing imaginary components were discarded, defining the reachable workspace. The resulting workspace (Fig.~\ref{fig:Workspace}) demonstrates that the mechanism can achieve continuous rotation about the vertical (yaw) axis, maintaining an identical workspace in the pitch and roll plane for every yaw angle. The reachable roll and pitch angles for each yaw angle form a spherical triangle, as shown in Fig.~\ref{fig:Workspace}. Thus, they will encompass a circular dome when considering all possible yaw angles. The dimensions of the spherical triangle in the roll and pitch plane are determined by the geometry of the arms and their mechanical interference, which limits the range of motion of the end-effector. 

Reviewing the experimental data presented in Fig.~\ref{fig:ConfigurationSpace} and Fig.~\ref{fig:Workspace} reveals noticeable discrepancies between the model and the empirical data. Instances where the experimental data exceed the theoretical boundaries arise from the exclusion of viable solutions due to the modelling of link interference in the kinematic analysis. Additionally, variations observed between the recorded workspace and the configuration space are due to the methodology used to evaluate the workspace. The workspace was assessed by back-driving the interface, which did not allow exploration of all possible configurations. Nevertheless, the symmetric behavior of the interface confirms that the proposed device is capable of covering the minimum desired workspace determined by the kinematic model, thereby satisfying the constraints required for the selected task described in Fig.~\ref{fig:UltrasoundProbeWorkspace}.

\section{Discussion}
The results demonstrate that the proposed parametric method enables adapting the interface to meet the requirements of different tasks and tools by optimising the structural parameters of the mechanism. In particular, larger tools necessitate an expanded workspace radius, while reduced angular motion requirements result in more compact, feasible designs. Such an approach facilitates the systematic evaluation of these trade-offs, supporting the design of compact, collision-free robotic structures tailored to specific clinical applications.

The FEA analysis confirmed that an aluminium prototype will allow it to support the required task loads with negligible deformations, adding slightly over 500 grams at the robotic end-effector. The tighter manufacturing tolerances and greater rigidity are likely to improve the workspace coverage for roll and pitch compared to the current PLA prototype manufactured using a Fused Deposition Modelling (FDM) 3D printer. Nevertheless, the current prototype was capable of smooth and continuous pure rotational motion throughout its available workspace.
\begin{figure}[htbp]
\centering
\includegraphics[width=\columnwidth]{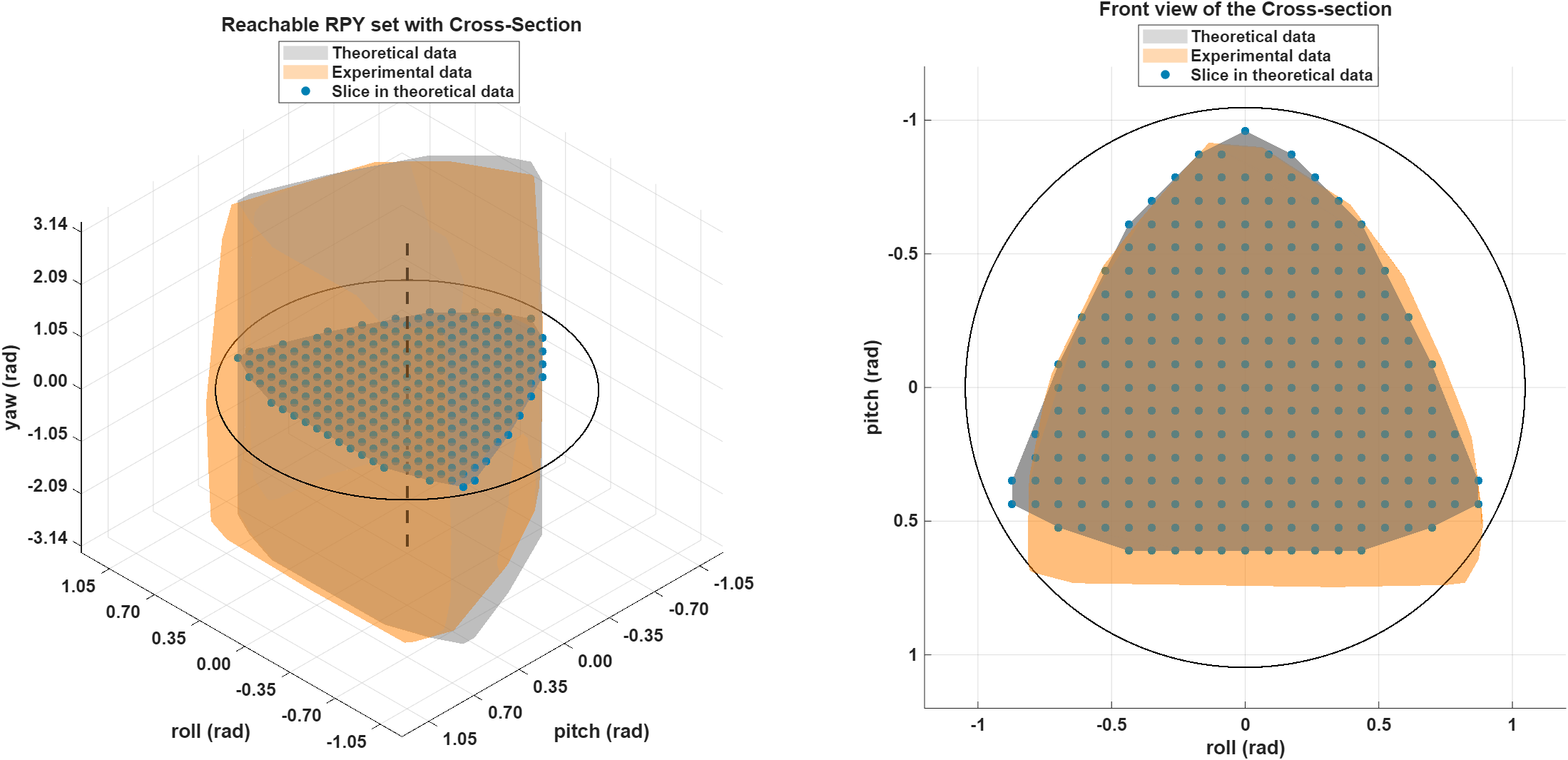}
\caption{
    Numerically computed workspace of CDC-SPM (gray) compared with experimentally measured (orange).
    }
\label{fig:Workspace}
\end{figure} 

The configuration space analysis reveals an important kinematic property of the CDC-SPM. The existence of an invariant cross-sectional profile along the diagonal direction of the joint space confirms that the mechanism inherently supports continuous rotation about the yaw axis. This property is particularly advantageous for applications such as ultrasound scanning, where uninterrupted rotational motion is required. However, it has shown that our theoretical interference constraints are not entirely accurate, demonstrating that the prototype can reach beyond our theoretical limits. Thus, it is essential to calibrate them on the physical prototypes to ensure that the model's output fully captures the interface's capabilities.

The workspace evaluation (Fig.~\ref{fig:Workspace}) further reiterates the need to perform a systematic evaluation by fully exploring the configuration space to ensure its completeness. Fig.~\ref{fig:ConfigurationSpace} shows that although the experiment covered most of the workspace, it did not fully explore the interface configuration space. However, the workspace (Fig.~\ref{fig:Workspace}) achieved with the current prototype will still be sufficient to perform an ultrasound scan as described in Fig.~\ref{fig:UltrasoundProbeWorkspace}.

\section{Conclusions}
The proposed mechanism is a viable solution to address the trade-off imposed by the manipulability requirements and dynamic interaction performance during ultrasound scans. Its compact design and low inertia provide a solid foundation for developing a high-frequency buffer between the robotic arm and the patient. However, there are certain limitations of the current design that need to be addressed before deploying it on a robotic arm. Therefore, the next step will involve introducing a tensioning mechanism for the cable to maximise the interface's range of motion, integrating sensors to improve the computational accuracy of the mechanism's model, and manufacturing an aluminium prototype for better rigidity. These are all characteristics that are essential to accurately evaluate the dynamic performance of the haptic device.





\bibliographystyle{asmejour}   

\bibliography{references.bib} 



\end{document}